\documentclass{article}

 \usepackage[preprint,nonatbib]{nips_2018}

\usepackage[utf8]{inputenc} % allow utf-8 input
\usepackage[T1]{fontenc}    % use 8-bit T1 fonts
\usepackage{hyperref}       % hyperlinks
\usepackage{url}            % simple URL typesetting
\usepackage{booktabs}       % professional-quality tables
\usepackage{amsfonts}       % blackboard math symbols
\usepackage{nicefrac}       % compact symbols for 1/2, etc.
\usepackage{microtype}      % microtypography
\usepackage{tabularx}

\usepackage{optidef}
\usepackage[font=footnotesize,labelfont=bf]{caption}

\usepackage{multirow}
\usepackage{tabularx}
\usepackage{dsfont}
\usepackage{url}
\usepackage[tight]{subfigure}
\usepackage{color}
\usepackage{subfigure}
\usepackage{amsthm}
\usepackage[ruled,vlined]{algorithm2e}
\usepackage[export]{adjustbox}
\usepackage{wrapfig}

\theoremstyle{definition}

\theoremstyle{remark}

\usepackage{blindtext}
\usepackage{enumitem}
\usepackage{xcolor}

\title{Proximal Splitting Networks for Image Restoration}

\author{
  Raied Aljadaany ~~~Dipan K. Pal ~~~Marios Savvides \\
  Department of Electrical and Computer Engg.\\
  Carnegie Mellon University\\
  Pittsburgh, PA 15213 \\
  \texttt{\{raljadaa, dipanp, marioss\}@andrew.cmu.edu} \\
}

\begin{document}

\maketitle

\begin{abstract}
% Our approach is a generalization of many previous methods.
Image restoration problems are typically ill-posed requiring the design of suitable priors. These priors are typically hand-designed and are fully instantiated throughout the process. In this paper, we introduce a novel framework for handling inverse problems related to image restoration based on elements from the half quadratic splitting method and proximal operators. Modeling the proximal operator as a convolutional network, we defined an implicit prior on the image space as a \textit{function class} during training. This is in contrast to the common practice in literature of having the prior to be fixed and fully instantiated even during training stages. Further, we allow this proximal operator to be tuned differently for each iteration which greatly increases modeling capacity and allows us to reduce the number of iterations by an order of magnitude as compared to other approaches. Our final network is an end-to-end one whose run time matches the previous fastest algorithms while outperforming them in recovery fidelity on two image restoration tasks. Indeed, we find our approach achieves state-of-the-art results on benchmarks in image denoising and image super resolution while recovering more complex and finer details.

\end{abstract}
%[BACKUP]Image restoration problems are typically ill-posed which require the design of suitable priors. These priors are typically hand-designed and are fully instantiated throughout the process. In this paper, we introduce a new framework for handling inverse problems related to image restoration based on elements from the half quadratic splitting methods and proximal operators. Modeling the proximal operator as a convolutional network, leads to an implicit prior being defined on the image as a function class during training. This is in contrast to the common practice in literature of having the prior be fixed and fully instantiated even during training stages. We further allow this proximal operator to be tuned differently for each iteration which greatly increases modeling capacity and allows for drastically reduced number of iterations. Our final network is an end-to-end one which runs matches the previous fastest algorithms while outperforming them. Indeed, we find our approach achieves state-of-the-art results on benchmarks in image denoising and image super resolution.

\section{Introduction}
% HOW IS PSN K AND PSN U DIFFERENT. TALK ABOUT THAT.

%ADD IN THAT WE REQUIRE LESS ITERATIONS
%

%\begin{enumerate}
   % \item we are selling strongly that our approach learns the prior, are there any other approaches that do the same? if yes, it is not clear in the text, we should clarify this
%\end{enumerate}

%\section{Fast ADMM-net}
%Several approaches have been proposed to solve equation \ref{eq:2}[add references]. One of theses approaches is ADMM algorithm [admm ref]. To apply ADMM algorithm, \ref{eq:2} can be written as:
%\begin{equation}
%{x}^*,{z_i}^* = arg \min_{{x},{z_i}}\|{y} - {k} *{x}\|^2_2 + \Sigma_i g(z_i) \quad z_i=D_i x \quad \forall i
%\label{eq3}
%\end{equation}
%The augmented Lagrangian form of \ref{eq3} is:
%\begin{equation}
%L(x,z,u) = arg \min_{{x},{z_i},{u_i}}\|{y} -{k} *{x}\|^2_2 + \Sigma_i g(z_i)+u_i^T(z_i-D_i x)+\rho_i\|{z_i} - {D_i} {x}\|^2_2 
%\label{eq4}
%\end{equation}
%Where $u_i$ is Lagrangian multiplier and $\rho_i$ is plenty variable. ADMM algorithm will find the optimum solution by solving the next three optimization problem alternatively:

%\section{The Image De-blurring Problem(DONE)}

% For example when the blur kernel is low pass filter \cite{hsu2006design},the high frequency features of an image or a signal will be lost.
%\begin{equation}
%{x}^* = arg \min_{{x}}\|{y} - {k} *{x}\|^2_2 + \mu\|L {x}\|^2_2
%\label{eq:3}
%\end{equation}

Single image restoration aims to reconstruct a clear image from corrupted measurements. Assume a corrupted image $y$ can be generated via convolving a clear image $x$ with a known linear space-invariant blur kernel $k$. This can be written as:
\begin{equation}
y = k* x + \epsilon
\label{eq:1}
\end{equation}
where $\epsilon$ is an additive zero-mean white Gaussian noise and $*$ is the convolution operation. The problem of recovering the clean image is an ill-posed inverse problem. One approach to solve it is by assuming some prior (or a set of) on the image space. Thus, the clean image can be approximated by solving the following optimization problem
\begin{equation}
{x}^* = arg \min_{{x}}\|{y} - {k} *{x}\|^2_2 + g(x)
\label{eq:2}
\end{equation}
where $\|.\|_2$ is the $l_2$ norm and $g$ is an operator that defines some prior (e.g $l_1$ norm is used to promote sparsity). A good prior is important to recover a feasible and high-quality solution. Indeed, priors are  common in signal and image processing tasks such as inverse problems \cite{mallat2008wavelet,kim2010single} and these communities have spent considerable effort in hand designing suitable priors for signals \cite{antonini1992image,rubinstein2010dictionaries,sun2008image}. 

%For instance, there have been many approaches proposed which build the prior operator by transforming the image linearly and assuming the transformed image has small energy or sparse etc \cite{hoeher1997two,beck2009fast,altintacc1988}.

%A classical prior is to transform the image by using the Laplacian matrix (filter) and assume the transformed image has minimum energy. This prior is well-known  to be related to Wiener filtering \cite{hoeher1997two}.

% The network on the other hand, can model a very large class of functions according to the universal approximation theory \cite{chen1995universal}.
%SAY PRIOR IS MORE FLEXIBLE, THE STRUCTURE OF THE FUNCTION INDUCES A PRIOR

In this work, we build a framework where the image prior is a \textit{function class} during training, rather than a specific instantiation. Parameters of this function are learned which then acts as a fully instantiated prior during testing. This allows for a far more flexible prior which is learned and tuned according to the data. This is somewhat in contrast to what the concept of a prior is in the general machine learning setting where it is usually a specific function (rather a function class). In our application for image restoration, the prior function class is defined to be a deep convolutional network. Such networks have exceptional capacity at modelling complex functions while capturing natural structures in images such as spatial reciprocity through convolutions.  A large function class for the prior allows the optimization to model richer statistics within the image space, which leads to better reconstruction performance (as we find in our experiments).

%The inherent inductive bias present in the network acts as prior knowledge.

%Which means that the network will learn to transform the image to another domine then the network will learn which constraints must be applied to recover the clear image without making any assumption about the data.

%Thus, we will use a deep network to find the prior function of the data.
%, neural network has the ability to represent any function belongs to Hilbert space\cite{gleason1957measures}.
% If known, here is an additional option of providing the noise level for denoising applications. Nonetheless, our algorithm performs well even if the level is unknown.

Our reconstruction network  takes only two inputs, the corrupted image and the kernel, then it reconstructs the clear image with a single forward pass. The network architecture is designed following a recovery algorithm based on the half quadratic splitting method involving the proximal operator (see section 3).  Proximal operators have been successfully applied in many image processing tasks (\emph{e.g.} \cite{nikolova2005analysis}). They are powerful and can work under more general conditions (e.g don't require differentiability) and are simple to implement and study. Theoretically, the reconstruction process is driven primarily by the half-quadratic splitting method with no trainable parameters. The only need for training arises when the proximal operators in this architecture are modelled using deep networks. This training only helps the network to learn parameters such that the overall pipeline is effective. Our overall framework is flexible and can be applied to almost any image based inverse problem, though in this work we focus on image denoising and image super-resolution. 

%We find in our experiments, that the networks out-performs state-of-the-art algorithms both on image denoising and image super resolution tasks.

\textbf{Contributions.}
%\begin{description}[font=$\bullet$~\normalfont\scshape\color{red!50!black}] called Proximal Splitting Networks (PSN)
 We propose a novel framework for image restoration tasks called Proximal Splitting Networks (PSN). Our network architecture  is theoretically motivated through the half quadratic splitting method and the proximal operator. Further, we model the proximal operators as deep convolutional networks that result in  more flexible signal priors tuned to the data, while requiring a number of iterations an order of magnitude less compared to some previous works. Finally, we demonstrate state-of-the-art results on two image restoration tasks, namely image denoising and image super resolution on multiple standard benchmarks. %Lastly, we show that our approach is a generalization of other well-known approaches in image recovery utilizing deep learning (in the supplementary).
%\item  We show that the structure of the proposed framework is  motivated by a theoretical background that is based on the proximal operator and half quadratic split method

%\end{description}

\begin{figure}
\centering
  \includegraphics[width=\linewidth]{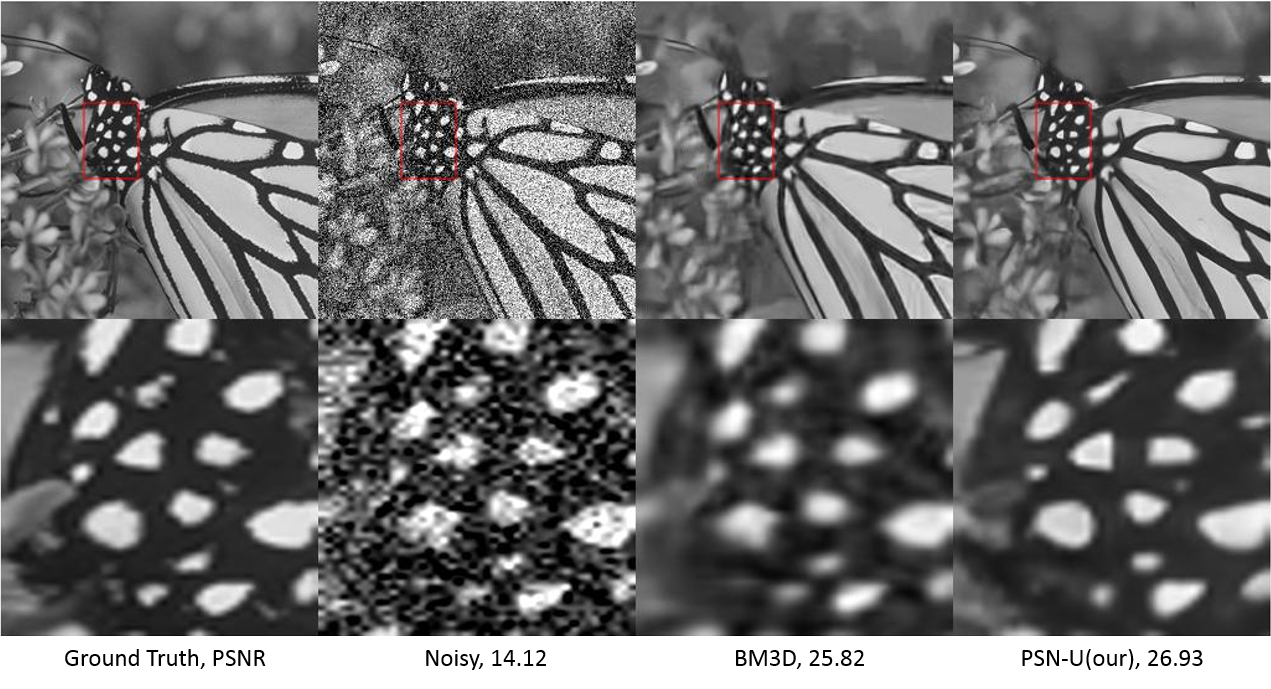}
  \caption{De-noising results of  “Fly” (Set12) with Gaussian noise added with $\sigma=50$. Numbers indicate PSNR. Our algorithm called Proximal-Splitting Network (PSN, U denotes unknown noise level) produces sharper edges in various orientations, an artifact of a powerful learned prior. For illustration, we compare against BM3D \cite{dabov2007image}.}
  \label{fig_Fly}
%%\vspace{-0.5cm}
\end{figure}

%%%%%%%%%%%%%%%%%%%%%%%%%%%%%%%%%%%%%%%%%%%%%%%%%%%%%%%%%%%%%%%%%%%%%%%%%%%%%%%%%%%%%%%%%%%%%
%%%%%%%%%%%%%%%%%%%%%%%%%%%%%%%%%%%%%%%%%%%%%%%%%%%%%%%%%%%%%%%%%%%%%%%%%%%%%%%%%%%%%%%%%%%%%
%%%%%%%%%%%%%%%%%%%%%%%%%%%%%%%%%%%%%%%%%%%%%%%%%%%%%%%%%%%%%%%%%%%%%%%%%%%%%%%%%%%%%%%%%%%%%

%Under this setting, Eq.~\ref{eq:2} will become ${x}^* = arg \min_{{x}}\|{y} - {k} *{x}\|^2_2 + \mu\|L {x}\|^2_2$. A popular traditional prior is to transform the image by using the Laplacian matrix (filter) and assume the transformed image has minimum energy. This prior is well-known  to be related to Wiener filtering \cite{hoeher1997two}. However, this prior forces the image to be smooth which cause blurring of the image when used image enhancement tasks.
\section{Prior Art}

\textbf{Priors in image restoration.}  The design of priors for solving inverse problems has enjoyed a rich history. There have been linear transforms proposed as priors which also assume low energy or sparsity etc \cite{hoeher1997two,beck2009fast,altintacc1988}. However, it has been shown that these approaches  fail when the solution space invoked by the assumed prior does not contain good approximations of the real data \cite{elad2006image}. There have also been other signal processing methods such as BM3D \cite{dabov2007image}, and those levaraging total variation \cite{wang2008new} and dictionary learning  \cite{rubinstein2010dictionaries,elad2006image,tosic2011dictionary} techniques that have been successful in several of these tasks. Proximal operators \cite{parikh2014proximal} and half quadratic splitting \cite{wang2008new} methods have been useful in a few image recovering algorithms such as \cite{beck2009fast}. These methods assume hand-designed priors approximated via careful choice of the norm. Although this has provided much success, they are limited in the expressive capacity of the prior which ultimately limits the quality of the solution. In our approach, we \textit{learn} the prior from a function class for our problem during the optimization. Thus our algorithm utilizes a more expressive prior that is informed by data.

%To combat this issue, a few approaches have been proposed to compute the prior from the data such as principle Component Analysis (PCA) \cite{wold1987principal} and dictionary learning \cite{rubinstein2010dictionaries}. However, these data driven approaches is built on making some assumption on the prior. For instance, low-rank PCA projection transforms the image non-linearly and assumes that transformed image is low rank. Dictionary learning techniques assume a similar but slightly more expressive prior. In dictionary learning, the transformed image is assumed to be sparse with respect to the dictionary basis. These approaches thus have limitations and are constrained in the expressive capacity of the prior. This limitation seems to be a major bottleneck towards achieving higher reconstruction performance.

\textbf{Deep learning approaches and our generalization.}  Deep learning approaches have emerged successful in modelling high level image statistics and thus have excelled at image restoration problems. Some example applications include blind de-convolution \cite{nah2017deep}, super-resolution \cite{kim2016accurate} and  de-noising\cite{zhang2017beyond}. Though these methods have powerful generalization, it remains unclear as to what the relation between the architecture and the prior used is. In this work however, the network is clearly motivated based on a combination of the proximal operator and half quadratic splitting methods. Further, we show that our network is a generalization of the approaches in \cite{kim2016accurate,zhang2017beyond} in the supplementary.

\begin{figure}
  \includegraphics[width=\linewidth]{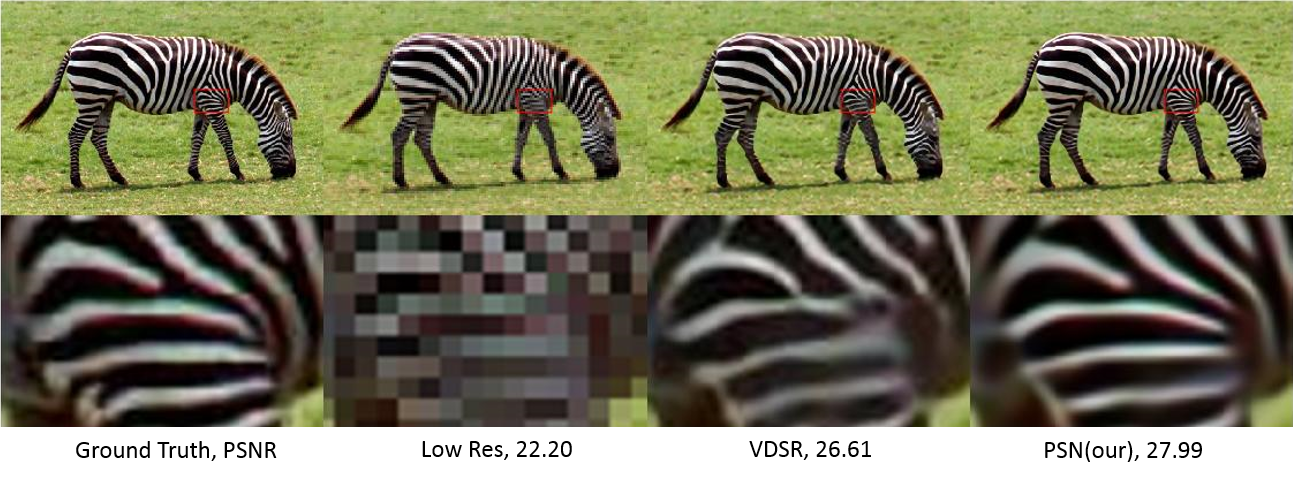}
  \caption{Super-resolution results of “Zebra”(Set14) downsampling (scale) factor X4.  Numbers indicate PSNR. Proximal-Splitting Network (PSN) produces much sharper and well-defined edges in various orientations, as artifact of a powerful learned prior. For illustration, we compare against VDSR \cite{kim2016accurate}.}
  \label{fig_s5}
%%\vspace{-0.5cm}
\end{figure}

\textbf{Deep learning approaches which learn the prior} It is worth mentioning that several approaches have used a proximal gradient decent algorithm \cite{meinhardt2017learning}, ADMM algorithm \cite{rick2017one} or a gradient decent method \cite{bigdeli2017deep} to recover an image where the prior is computed via a deep learning network. Although, these approaches preform well with respect to the reconstruction performance, they inherit important limitations in terms of computation efficiency. Proximal gradient decent, ADMM and gradient decent methods being first order iterative methods with linear or sub-linear convergence rate, typically require many tens of iterations for convergence. Each iteration consists of a forward pass through the network, which emerges as a considerable bottleneck. Our approach addresses this problem by allowing different proximal operators to be learned at every `iteration'. This increases modelling capacity of the overall network and allows for much lower iterations (an order of magnitude lower in our case).

\textbf{Deep learning structure based on theoretical approaches} There are several approaches that employed CNNs for image restoration \cite{schmidt2014shrinkage,jin2017noise,chen2017trainable} where the structure of the network is driven from a theoretical model for image recovery. In \cite{schmidt2014shrinkage}, the author proposed the cascade of shrinkage fields for image restoration. CSF can be seen as a ConvNet where the architecture of this network is a cascade of Gaussian conditional random fields \cite{schmidt2013discriminative}.  \cite{chen2017trainable} proposed trainable nonlinear reaction diffusion (TNRD) which is a ConvNet that has structure based on nonlinear diffusion models \cite{perona1990scale}. In \cite{jin2017noise}, the authors proposed GradNet applied to noise-blind deblurring. The architecture of GradNet is motivated by the Majorization-Minimization (MM) algorithm \cite{hunter2004tutorial}. However, these approaches assume that the prior term is driven from or approximated by Gaussian mixture model \cite{zoran2011learning} which is represented by ConvNet. Our method is free of this assumption.

\section{Proximal Splitting in Deep Networks}

Our main goal is the design of a feed forward neural network for image restoration. For the architecture, we take inspiration from two tools in optimization. The first being the proximal operator which allows for a solution to a problem to be part of some predefined solution space. The second component being the half quadratic splitting technique which allows a sum of objective functions to be solved in alternating sequence using proximal operators. We briefly describe these two components and then utilize them to design our system architecture.

\begin{figure}
  \includegraphics[width=\linewidth]{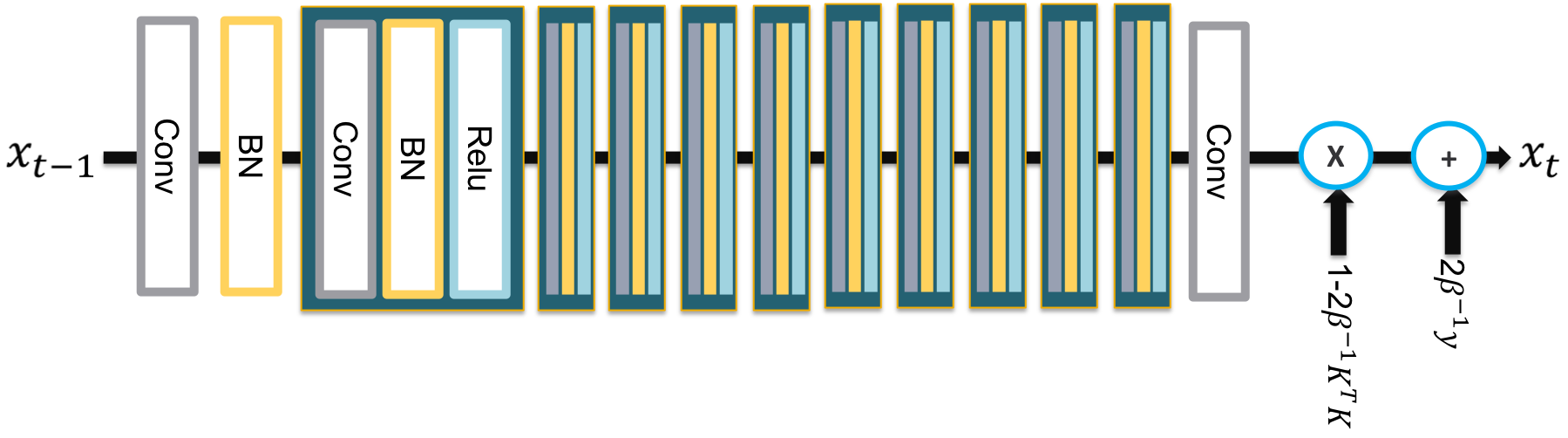}
  \caption{ Architecture of a single Proximal Block, the fundamental component of the Proximal Splitting Network (PSN). This block implements a single iteration of the half quadratic splitting method for the constraints in PSN optimization problem (Eq.~\ref{eq_newprob}). The block has 10 layers, each being the Convolution, ReLU and Batch Normalization \cite{ioffe2015batch} combination. All intermediate layers had 64 channels with $3\times 3$ convolutions, except the first/last Conv layer with 3/1 channel (RGB /grey-scale).}
  \label{fig_prox_block}
  %\vspace{-0.5cm}
\end{figure}

\subsection{Proximal Operator}

%Where $f(x)$ is convex and differential function and $g(x)$ can be any convex function. if $g(x)$ is not convex, the proximal gradient decent algorithm will converge to local minimum solution. Moreover,  If the proximal gradient algorithm is applied to  solve \ref{eq4}, the updating step at $k$ iteration will be:
%\begin{equation}
%{x}^{(k)} = prox_{gt_k}({x}^{(k-1)}-t_k \nabla f(x^{(k-1)}))
%\label{eq:5}
%\end{equation}
%Where $t_k$  is the step size at kth iteration. If $\nabla f$ belongs to Lipschitz space with constant $L$ and the step size is chosen to be between zero and $1/L$, the proximal gradient algorithm will converge to the exact solution with a linear convergence rate $O(1/k)${\cite{parikh2014proximal}}.

Let $h :R^n \rightarrow R $ be a function. The proximal operator of the function $h$ with the parameter $\beta$ is defined as
\begin{equation}
prox_{h,\beta}(x) = arg  \min_{{z}}\beta\|{z} - {x}\|^2_2 +h(z)
\label{eq:6}
\end{equation}
% This operator is a generalization of the projection operator. 
 If the function $h(x)$ is a strong convex function and twice differentiable with respect to $x$ and $\beta$ is large,  the proximal operator of the function $h$ converges to a gradient descent step (a proof of this known result is presented in the supplementary). In this case the proximal operator can be approximated as:
\begin{equation}
prox_{h,\beta}(x) \approx x-2\beta^{-1}\nabla h(x)
\label{eq:P_to_gradint}
\end{equation}

%There are several approaches have been proposed to find  the prior knowledge of the data. These approaches can be dived into two main category. The first approach is built on making assumptions on the signal or the data. For example, the data is sparse with respect wavelet transform [add several references]. Another method that is used to find the prior is by learning the prior from the data[.....].  

\subsection{Half Quadratic Splitting}

Now note that the image recovery optimization problem (Eq.~\ref{eq:2}) can be rewritten as:
\begin{equation}
{x}^* = arg \min_{{x}}f(x)+g(x)
\label{eq:4}
\end{equation}
where $f$ is the data fidelity term and $g$ is a function that represents the prior. Depending on this prior $g(x)$, Eq.~\ref{eq:4} might be hard to optimize, especially when the prior function is not convex. The half quadratic splitting method \cite{wang2008new} restructures this problem  (Eq.~\ref{eq:4}) into a constrained optimization problem  by introducing an auxiliary variable $v$. Under this approach, the optimization problem in Eq.~\ref{eq:4} is reformulated as: 
\begin{equation}
{x}^*,v^* = arg \min_{{x},v}f(x)+g(v),  \; s.t. \;  v=x
\label{eq:7}
\end{equation}
The next step is to convert the equality constraint into its Lagrangian.
\begin{equation}
{x}^*,v^* = arg \min_{{x},v}f(x)+g(v)+\beta \|{v} - {x}\|^2_2
\label{eq:8}
\end{equation}
where $\beta$ is a penalty parameter. As $\beta$ approaches infinity, the solution of Eq.~\ref{eq:8} is equivalent to that of Eq.~\ref{eq:4}, and can be solved in iterative fashion by fixing one variable, updating the other and vice versa. By using the proximal operator, these updating steps become
\begin{equation}
\begin{aligned}
{x}_t =prox_{f,\beta}(v_{t})~~~ v_t = prox_{g,\beta}(x_{t-1})
\end{aligned}
\label{eq:prox_update}
\end{equation}
%This approach will lead to the a global optimum solution of $x$ if the optimization problem is convex or to a local minimum solution if it is not(ADD REFER). 
When the image $x$ is fixed, the optimum $v$ can be found through the proximal operator involving $g(z)$ and $\beta$.
%\begin{equation}
%v_t=prox_{g,\beta}(x_{t}) 
%\label{eq:10}
%\end{equation}
Clearly, this depends on the prior which is the $g$ function. For instance, if $g(z)$ is $l_1$ norm, the prox operator will be a soft threshold operator which forces the signal to be sparse \cite{beck2009fast}. However, for real-world image data, the optimal class of functions for $g$ is not known, which by extension makes the prox-operator sub-optimal for recovery. In the following subsection, we will propose an approach to optimize for the prox operator within a predefined search space.

%FLOW BREAK HERE (RAIED)
As a final note, recall that since the added noise is assumed to be Gaussian, $f$ is the euclidean distances between the corrupted image and the clean image convolved with  a kernel. Thus, $f(x)=\frac{1}{2}\|{y} - {k} *{x}\|^2_2$ which is convex and twice differentiable.  This allows the updating step in Eq.~\ref{eq:prox_update} for $x$ to be approximated via gradient decent while modifying the proximal operator from Eq.~\ref{eq:P_to_gradint}:
 \begin{equation}
x_t=v_{t}-2\beta^{-1}[K^T(Kv_{t}-y)] 
\label{eq:11}
\end{equation}
 where   $K$ is the matrix form of the convolution operation with $k$ and $K^T$ is its transpose.

%\begin{equation}
%x_t=v_{t}-\beta[K^T(Kx_{t-1}-y)+\beta(x_{t-1}-v_t)] 
%\label{eq:11}
%\end{equation}

 %\begin{figure}%{r}{0.50\textwidth}
%\begin{figure}%{r}{0.5\textwidth}
  % \begin{center}
 %       \subfigure[Proximal Splitting Network]{%
%        \centering
       %     \includegraphics[width=0.45\columnwidth,valign=m]{network.png}\label{fig_1_a}
      %  }%
     %    \subfigure[Architecture of the Proximal Block of PSN]{%
    %    \centering
   %     \includegraphics[width=0.45\columnwidth,valign=m]{prox_block_i.png}\label{fig_1_b}
  %      }
 %   \end{center}
%\centering
%\includegraphics[width=0.45\columnwidth,valign=m]{network.png}
%    %\vspace{-0.0cm}
%\caption{ Architecture of (a) the overall Proximal Splitting Network (PSN) and (b) a single Proximal Block.   } 
%\label{fig_net}
%%\vspace{-0.5cm}
%\end{figure}

 \begin{figure}
  \includegraphics[width=0.9\linewidth]{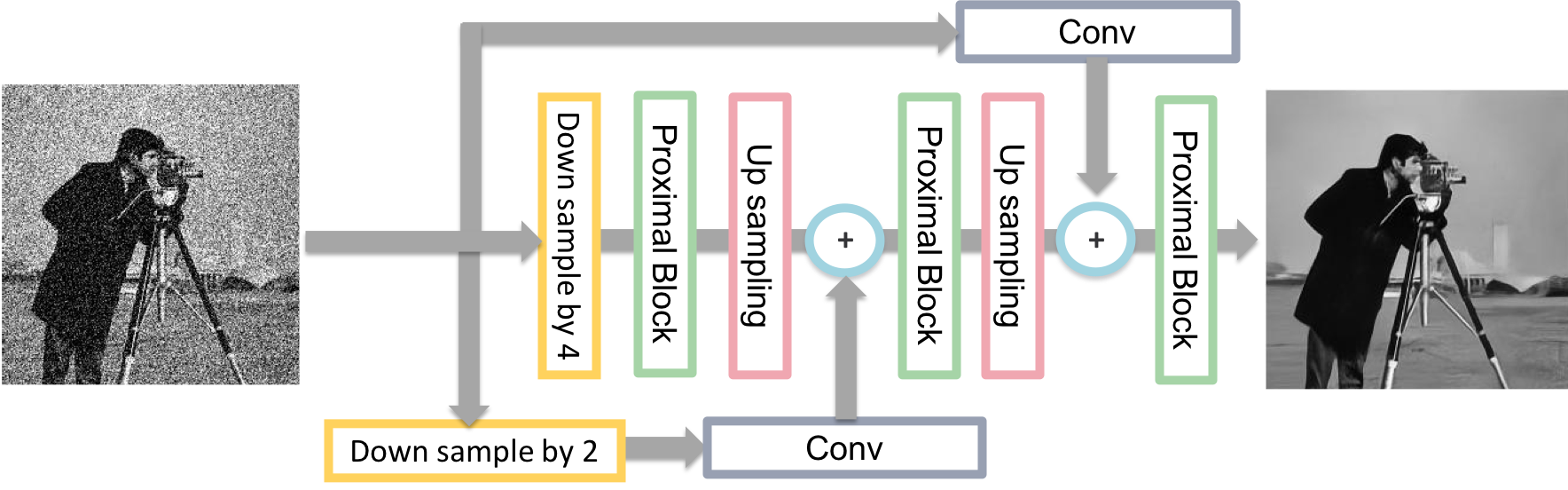}
  \caption{ The multi-scale Proximal Splitting Network (PSN) architecture for image restoration tasks. Each Up-sampling block does so by a factor of 2 whereas each Proximal Block has 10 layers (see Fig.~\ref{fig_prox_block}). The two Conv layers have 1 or 3 channels each (Grey-scale vs. RGB image space) with $3\times 3$ kernels.}
  \label{fig_network}
  %\vspace{-0.5cm}
\end{figure}
 
 % with 3 stages/iterations (3 Proximal Blocks)

\subsection{Proximal Splitting Networks}

 We now develop the core optimization problem which will then yield the Proximal Splitting Network architecture. Our main approach for image recovery is to use the half quadratic splitting method which alternately updates the image $x$ and an auxiliary variable $v$ as in Eq.~\ref{eq:prox_update}. Thus, for $S$ iterations the optimization procedure becomes
 \begin{equation}
%\begin{aligned}
  v_t = prox_{g,\beta}(x_{t-1}), ~~ x_t=v_{t}-2\beta^{-1}[K^T(Kv_{t}-y)]~~ \forall t = 1, \ldots, S
%\end{aligned}
\label{eq_oriprob}
\end{equation}
Note that the update for $v_t$ still contains a proximal operator depending on  the prior $g$. There have been studies such as \cite{meinhardt2017learning}, where the authors replace the proximal operator with a Deep Denoising Network (DnCNN) \cite{zhang2017beyond}. Similarly, the authors in \cite{heide2014flexisp}  use BM3D or the NLM denoiser rather than the proximal operator to update the value of an image. It is also important to note that these studies utilized these denoisers in an iterative fashion \emph{i.e.} the \textit{same} proximal operator with its parameters was used through multiple iterations.  Considering that the number of iterations in these studies were significantly high (about 30 for both \cite{meinhardt2017learning} and \cite{heide2014flexisp}) and the fact that every iteration requires a forward pass through a deep network, these methods have large computational bottleneck.

Although these methods work well, there is much to gain from defining a more flexible proximal operator in two ways. First, defining a larger solution space for the proximal operator would allow for the algorithm to choose more fitting operators. Secondly, allowing the proximal operator networks at different stages (iterations) to maintain separate weights allows for each operator to be tuned to the statistics of the estimated image at that stage. This also allows us to keep the number of iterations or stages very small in comparison due to the larger modelling capacity (3 in our experiments, which is an order of magnitude less than previous studies  \cite{meinhardt2017learning,heide2014flexisp}). Keeping these in mind, we choose the model for the proximal operator in our formulation to be a deep convolutional network, which introduces desirable inductive biases. These biases themselves act as our `prior' while providing the optimization a large enough function search space to choose from. The rest of the prior (\emph{i.e.} the actual parameters of the convolutional network) are tuned according to the data. Under this modification, the update step for $v_t$ becomes  \begin{equation}
v_t=\Gamma^t_\Theta(x_{t-1})
\label{eq:10}
\end{equation}
where $\Gamma^t_\Theta$ is a convolutional network for the $t^{th}$ iteration. Note that for every iteration, there is a separate such network. Defining the proximal operator (and the image prior) to be different for every iteration, the final optimization problem becomes
\begin{equation}
\begin{aligned}
 \min_{{\Theta}}L({x_{gt}},x_{S})~~~ \text{s.t}~~~  v_t=\Gamma^t_\Theta(x_{t-1}), ~~ x_t=v_{t}-2\beta^{-1}[K^T(Kv_{t}-y)]  \; \forall t = 1, \ldots, S
\end{aligned}
\label{eq_newprob}
\end{equation}
 where $x_{gt}$ is the ground truth clean image, $x_S$ is the final estimated image, $S$ is the number of stages (iterations) and $x^0$ is the initial input image. Note that the minimization in this formulation is only on $\Theta$ \emph{i.e.} the parameters of the set of proximal networks $\Gamma^t_\Theta ~~\forall t$. The loss function here can be any suitable function, though we minimize the Euclidean error for this study assuming Gaussian noise. It is important to note a subtle point regarding the recovery framework. The minimization in Eq.~\ref{eq_newprob} only tunes the network $\Gamma_\Theta$ towards the desired task based off the data. However, the core algorithm for reconstruction is still based on Eq.~\ref{eq_oriprob} \emph{i.e.} iterations of the half quadratic splitting based reconstruction.   It is also useful to observe the interplay between the objective function and the constraints. The first constraint and the loss objective in Eq.~\ref{eq_newprob} work to project the recovered image onto the image space  while the second constraint pushes the recovered image to be as close as possible to the the corrupted input image. A single iteration over these constraints according to half quadratic splitting, and the proximal network $\Gamma_\Theta$ together result in what we call the Proximal Block as shown in Fig.~\ref{fig_prox_block}. The Proximal Block is the fundamental component using which the overall network is built (as we describe soon).

\textbf{Multi-scale Proximal Splitting Network.} Multi-scale decomposition has been widely applied in many applications, such as edge-aware filtering \cite{paris2011local}, image blending \cite{burt1987laplacian} and semantic segmentation \cite{ghiasi2016laplacian}.  Multi-scale architecture extensions have also emerged as a standard technique to further improve the performance of deep learning approaches to image recovery tasks such as image de-convolution \cite{nah2017deep} and image super-resolution \cite{lai2017deep}. We find that the multi-scaling is useful incorporate it into the Proximal Splitting Network algorithm. These approaches usually require that the output of each intermediate scale stage be the cleaned/processed image at that scale. Complying with this, multi-scaled PSN networks are designed such that the intermediate outputs form a Gaussian pyramid of cleaned images. For better performance, we apply reconstruction fidelity loss functions at each level of the pyramid. This also helps provide stronger gradients for the entire network pipeline, especially the first few layers which are typically harder to train. 

\textbf{Proximal Splitting Network Architecture for Image Restoration.} Finally, we implement Eq.~\ref{eq_newprob} to arrive at the PSN architecture while utilizing multiple Proximal Blocks in Fig.~\ref{fig_network}. The number of Proximal Blocks (from Fig.~\ref{fig_prox_block}) equals the number of stages or iterations for the half-quadratic splitting method (Eq.~\ref{eq_newprob}) which we set to be 3 \emph{i.e.} $S=3$. Recall that this is an order of magnitude lesser than some previous works \cite{meinhardt2017learning,heide2014flexisp}.  In Fig.~\ref{fig_network}, the input image is the corrupted image convolved with the $K^T$(e.g the input image is the noisy image for image denoising and it is the up sampled image via bi cubic interpolation for image super-resolution in the experiment part) . The down sampling is achieved via bi-cubic down sampling and the up sampling by a de-convolution layer \cite{noh2015learning}. Through a preliminary grid search, we find that $\beta=8$ works satisfactorily. 
%\subsection{Loss function}

\begin{table}
\small
\begin{tabularx}{\textwidth}{ |c|X|X|X|X|X|X|X|X|X| }
  \hline
  $\sigma$ & BM3D \cite{dabov2007image} & WNNM \cite{gu2014weighted} & EPLL \cite{zoran2011learning} & MLP \cite{burger2012image} & CSF \cite{schmidt2014shrinkage} & TNRD \cite{chen2017trainable} & DnCNN \cite{zhang2017beyond} & PSN-K  (Ours)  & PSN-U (Ours) \\
  \hline
   15 & 31.07 & 31.37 & 31.21 & --- & 31.24 & 31.42 & \textit{31.61} & \textbf{31.70} & \textit{31.60}\\
  \hline
   25 & 28.57 & 28.83 & 28.68 & 28.96 & 28.74 & 28.92 & \textit{29.16} & \textbf{29.27} & \textit{29.17}  \\
      \hline
       50 & 25.62 & 25.87 & 25.67 & 26.03 & --- & 25.97 & 26.23 & \textbf{26.32} & \textit{26.30}\\
      \hline

\end{tabularx}
\caption{Denoising PSNR test results of several algorithms on BSD68 with noise levels of $\sigma=\{15,25,50\}$. \textbf{Bold} numbers denote the highest performing model, whereas \textit{Italics} denotes the second highest. PSN outperforms previous state-of-the-art when the noise level is known, however matches it when it is unknown. }
\label{table:set68}
%\vspace{-0.2cm}
\end{table}

%%%%%%%%%%%%%%%%%%%%%%%%%%%%%%%%%%%%%%%%%%%%%%%%%%%%%%%%%%%%%%%%%%%%%%%%%%%%%%%%%%%%%%%%%%%%%
%%%%%%%%%%%%%%%%%%%%%%%%%%%%%%%%%%%%%%%%%%%%%%%%%%%%%%%%%%%%%%%%%%%%%%%%%%%%%%%%%%%%%%%%%%%%%
%%%%%%%%%%%%%%%%%%%%%%%%%%%%%%%%%%%%%%%%%%%%%%%%%%%%%%%%%%%%%%%%%%%%%%%%%%%%%%%%%%%%%%%%%%%%%

\section{Empirical Evaluation on Image Restoration}
%where the step size range is from $10^{-2}$ to $10^{-5}$
We evaluate our proposed approach against state-of-the-art algorithms on standard benchmarks for the tasks of image denoising and image super resolution. For training we use Adam \cite{kingma2014adam} for 50 epochs with a batch size of 128 for all models. Runtimes for evaluated PSN network are on par with the fastest algorithms while outperforming previous state-of-the-arts (provided in the supplementary).

%\footnote{We provide comparative computational run times in the supplementary.}

%\linespread{0.1}

\subsection{Image De-noising}

Our first task is image denoising where given a noisy image (with a known and unknown  level of noise), the task is to output a noiseless version of the image. Image denoising is considered as special case of Eq.~\ref{eq:1} where $k$ is a delta function with no shift.

\textbf{Experiment:} We train on 400 images of size $180\times 180 $ from the Berkeley Segmentation Dataset (BSD)  \cite{arbelaez2007berkeley}. We set the patch size as $64\times 64$, and crop about one million random patches to train. We train four models as described in \cite{zhang2017beyond}. Three of these models are trained on images with three different levels of Gaussian noise i.e., $\sigma$ = 15, 25 and 50. We refer to these models as PSN-K (Proximal Split Net-Known noise level). The fourth model is trained for blind Gaussian denoising, where no level of $\sigma$ is assumed. For blind Gaussian denoising, we train a single model and set the \textit{range} of the noise level in the training images to be $\sigma \in$ [0, 60]. We refer to these models as PSN-U (Unknown noise level). We test on two well known datasets, the Berkeley Segmentation Dataset (BSD68) \cite{roth2009fields} containing a total of 68 images and Set12 \cite{dabov2007image} with 12 images with no overlap during training. We compare our approach with several state-of-the-art methods such as BM3D \cite{dabov2007image}, WNNM \cite{gu2014weighted}, TRND \cite{chen2017trainable},  EPLL \cite{zoran2011learning}, DnCNN \cite{zhang2017beyond}, MLP \cite{burger2012image} and CSF \cite{schmidt2014shrinkage}.

\textbf{Results:}  Table~\ref{table:set68} showcases the testing PSNR results on BSD68. We observe that PSN-K outperforms all other algorithms to obtain a new state-of-the-art on BSD68. However, the noise-blind version (PSN-U) very closely matches the previous state-of-the-art and for $\sigma=25, 50$ outperform it. Table~\ref{table:set12} shows the testing PSNRs for Set12. We find that for most images, PSN-K achieves new state-of-the-arts. The noise-blind model PSN-U also beats the state-of-art on many images in some cases even PSN-K. PSN-U performs particularly well at high levels of noise \emph{i.e.} $\sigma=50$. Fig.~\ref{fig_Fly} and Fig.~\ref{fig_additional_results} present some qualitative results illustrating the high level of detail PSN recovers. More results are presented in the supplementary.

%%%%%%%%%%%%%%%%%%%%%%%%%%%%%%%%%%%%%%%%%%%%%%%%%%%%%%%%%%%%%%%%%%%%%%%%%%%%%%%%%%%%%%%%%%%%%%%%
%%%%%%%%%%%%%%%%%%%%%%%%%%%%%%%%%%%%%%%%%%%%%%%%%%%%%%%%%%%%%%%%%%%%%%%%%%%%%%%%%%%%%%%%%%%%%%%%
%\begin{tabularx}{17 cm}{ |X|X|X|X|X|X|X|X|X|X|X|X|X| }
\begin{table}
\small
\centering
%\tiny
%\begin{tabular}{l*{11}{c}r}
\begin{tabular}{l*{11}{p{0.6cm}}r}
  \hline
   & C.Man & House & Pepp &  Starf. & Fly & Airpl. & Parrot & Lena & Barb. & Boat & Man & Couple \\
     \hline
       &   &   &   &    &  &  $\sigma=$15  &   &   &   &   &   &  \\
             \hline
                  \hline
   BM3D \cite{dabov2007image} & 31.91 &  34.93 & 32.69 &  31.14 & 31.85 &  31.07 & 31.37 & 34.26 & \textit{33.10} & 32.13 & 31.92 & 32.10\\
     CSF \cite{schmidt2014shrinkage}& 31.95 & 34.39 & 32.85 &  31.55 &  32.33 & 31.33 & 31.37 & 34.06 & 31.92 & 32.01 & 32.08 & 31.98\\
    EPLL  \cite{zoran2011learning}& 31.85 & 34.17 & 32.64 &  31.13 &  32.10 & 31.19 & 31.42 & 33.93 & 31.38 & 31.93 & 32.00& 31.93\\
   WNNM \cite{gu2014weighted}& 32.17 & \textbf{35.13} & 32.99 &  31.82 &  32.71 & 31.39 & \textit{31.62} & 34.27 & \textbf{33.60} & 32.27 & 32.11 & 32.17\\
    TNRD \cite{chen2017trainable}& \textit{32.19} & 34.53 & 33.04 &  31.75 &  32.56 & 31.46 & \textit{31.63} & 34.24 & 32.13 & 32.14 & 32.23& 32.11\\
   DnCNN \cite{zhang2017beyond}& 32.10 & 34.93 & {33.15} &  \textit{32.02} &  \textit{32.94} & 31.56 & \textit{31.63} & \textit{34.56} & 32.09 & \textit{32.35} & \textbf{32.41} & \textit{32.41}\\
   \hline
   PSN-K (Ours) & \textbf{32.58}  & \textit{35.04} & \textbf{33.23} & \textbf{32.17}   & \textbf{33.11}  & \textbf{31.75}  & \textbf{31.89}  & \textbf{34.62}  & 32.64  & \textbf{32.52}  & \textit{32.39}  &\textbf{32.43}  \\
   PSN-U (Ours)  &  32.04 & 35.03  & \textit{33.21}  & 31.94   & \textit{32.93}  & \textit{31.61}  & \textit{31.62}  & \textit{34.56}  & 32.49  & \textit{32.41}  & 32.37  & \textbf{32.43} \\

  \hline
         &   &   &   &    &  & $\sigma=$25  &   &   &   &   &   &  \\
             \hline
                  \hline
   BM3D \cite{dabov2007image}& 29.47 &  32.99 & 30.29 &  28.57 & 29.32 &  28.49 & 28.97 & 32.03 &  \textit{30.73} & 29.88 & 29.59 & 29.70\\
     CSF \cite{schmidt2014shrinkage}& 29.51 & 32.41 & 30.32 &  28.87. &  29.69 & 28.80 &  28.91 & 31.87 & 28.99 & 29.75 & 29.68 & 29.50\\
   EPLL \cite{zoran2011learning} & 29.21 & 32.14 & 30.12 &  28.48 & 29.35 & 28.66 & 28.96 & 31.58 & 28.53 & 29.64 & 29.57 & 29.46\\
   WNNM \cite{gu2014weighted}& 29.63 & 33.22 & 30.55 &  29.09 & 29.98 & 28.81 & 29.13 & 32.24 & \textbf{31.28} & 29.98 & 29.74 & 29.80\\
    TNRD \cite{chen2017trainable}& 29.72 & {32.53}  &30.57 &  29.09 & 29.85  & 28.88 & 29.18 & 32.00 & 29.41  & 29.91  & 29.87 & 29.71\\
   DnCNN \cite{zhang2017beyond}& \textit{29.94} & 33.05 & 30.84 &  \textit{29.34} & \textit{30.25} & \textit{29.09} & \textit{29.35} & 32.42 & 29.69 & 30.20 & \textit{30.09} & 30.10  \\
   \hline
   PSN-K (Ours)   & \textbf{30.28}  & \textbf{33.26}  &   \textbf{31.01} & \textbf{29.57}   & \textbf{30.30}  & \textbf{29.28}  &  \textbf{29.38} &  \textbf{32.57} & 30.17  & \textbf{30.31}   &  \textbf{30.10} & \textbf{30.18} \\
   PSN-U (Ours)   & 29.79  & \textit{33.23}  & \textit{30.90}  & 29.30   & 30.17  &  29.06 & 29.25  & \textit{32.45}  & 29.94  &   \textit{30.25} & 30.05  & \textit{30.12}  \\
   \hline
            &   &   &   &    &  & $\sigma=$50  &   &   &   &   &   &  \\
             \hline
                  \hline
   BM3D \cite{dabov2007image}& 26.13 &  29.69 & 26.68 &  25.04 & 25.82 &  25.10 & 25.90 & 29.05 &  \textit{27.22} & 26.78 & 26.81 & 26.46\\
   MLP \cite{burger2012image}& 26.37 &  29.64 & 26.68 &  25.43 & 26.26 &  25.56 & 26.12 & 29.32 &25.24 & 27.03 & 27.07 & 26.67\\
   WNNM \cite{gu2014weighted}& 26.45 &  \textit{30.33} & 26.95 &  25.44 & 26.32 &  25.42 & 26.14 & 29.25 & \textbf{27.79} & 26.97 & 26.95 & 26.64\\
    TNRD \cite{chen2017trainable} & 26.62 &  29.48 & 27.10 &  25.42 & 26.31 &  25.59 & 26.16 & 28.93 &25.70 & 26.94 & 26.98 & 26.50\\
   DnCNN \cite{zhang2017beyond}& \textit{27.03} &  30.02 & \textit{27.39} &  \textit{25.72} & \textit{26.83} &  \textit{25.89} & {26.48} & \textit{29.38} &  26.38 & \textit{27.23} & \textbf{27.23} &26.09\\
   \hline
   PSN-K (Ours)   &   \textit{27.10}  &  \textbf{30.34} & \textit{27.40}  & \textbf{25.84} & \textbf{26.92} & \textbf{25.90}  & \textit{26.56}  & \textbf{29.54}  & 26.45  & 27.20  & \textit{27.21}  & \textbf{27.09} \\
   PSN-U (Ours)   &  \textbf{27.21} & \textit{30.21}  & \textbf{27.53}  &  25.63   & \textbf{26.93}  & \textit{}{25.89}  & \textbf{26.62}  & 29.54  &\textit{26.56}  & \textbf{27.27}  & \textbf{27.23}  & \textit{27.04} \\

  \hline

  \hline
\end{tabular}
%%\vspace{-0.5cm}
\caption{Denoising PSNR results of several algorithms on Set12 with noise levels of $\sigma=\{15,25,50\}$. \textbf{Bold} numbers denote the highest performing model, whereas \textit{Italics} denotes the second highest. PSN outperforms state-of-the-art for many images.}
\label{table:set12}
\end{table}

%\end{tabularx}
%%%%%%%%%%%%%%%%%%%%%%%%%%%%%%%%%%%%%%%%%%%%%%%%%%%%%%%%%%%%%%%%%%%%%%%%%%%%%%%%%%%%%%%%%%%%%%%%
%%%%%%%%%%%%%%%%%%%%%%%%%%%%%%%%%%%%%%%%%%%%%%%%%%%%%%%%%%%%%%%%%%%%%%%%%%%%%%%%%%%%%%%%%%%%%%%%

\subsection{Image Super-Resolution}
Our second task aims to reconstruct a high-resolution image from a single low-resolution
image . Image super-resolution is considered as special case of Eq.~\ref{eq:1} where $k$ is a bicubic down sampling filter with no added noise. %Thus $K^T$ in equation \ref{eq_newprob} is  up sampling bicubic interpolation. 

\textbf{Experiment:} For training, we use  DIV2K dataset. The dataset consists of 800 training images (2K resolution).The data set is augmented with random horizontal flips and $90^\circ$ rotations. We set the high resolution patch size to be $128\times 128$. The low res patches are generated via bicubic down sampling of the high resolution patches. We trained a single model for each of three different scales i.e.,  2X, 3X and 4X. We test  our algorithm on four benchmark datasets. The datasets are Set5 \cite{bevilacqua2012low}, Set14 \cite{zeyde2010single}, BSDS100 \cite{arbelaez2011contour} and URBAN100 \cite{huang2015single} . We compare our approach with several state-of-the-art methods such as A+ \cite{timofte2014a+}, RFL \cite{schulter2015fast}, SelfExSR \cite{huang2015single}, SRCNN \cite{dong2016image}, FSRCNN \cite{dong2016accelerating}, SCN \cite{wang2015deep}, DRCN \cite{kim2016deeply}, LapSRN \cite{lai2017deep} and VDSR \cite{kim2016accurate} in terms of the PSNR and SSIM metrics as in \cite{kim2016accurate}.

\begin{figure}[h]
 \centering
  \includegraphics[width=0.8\linewidth]{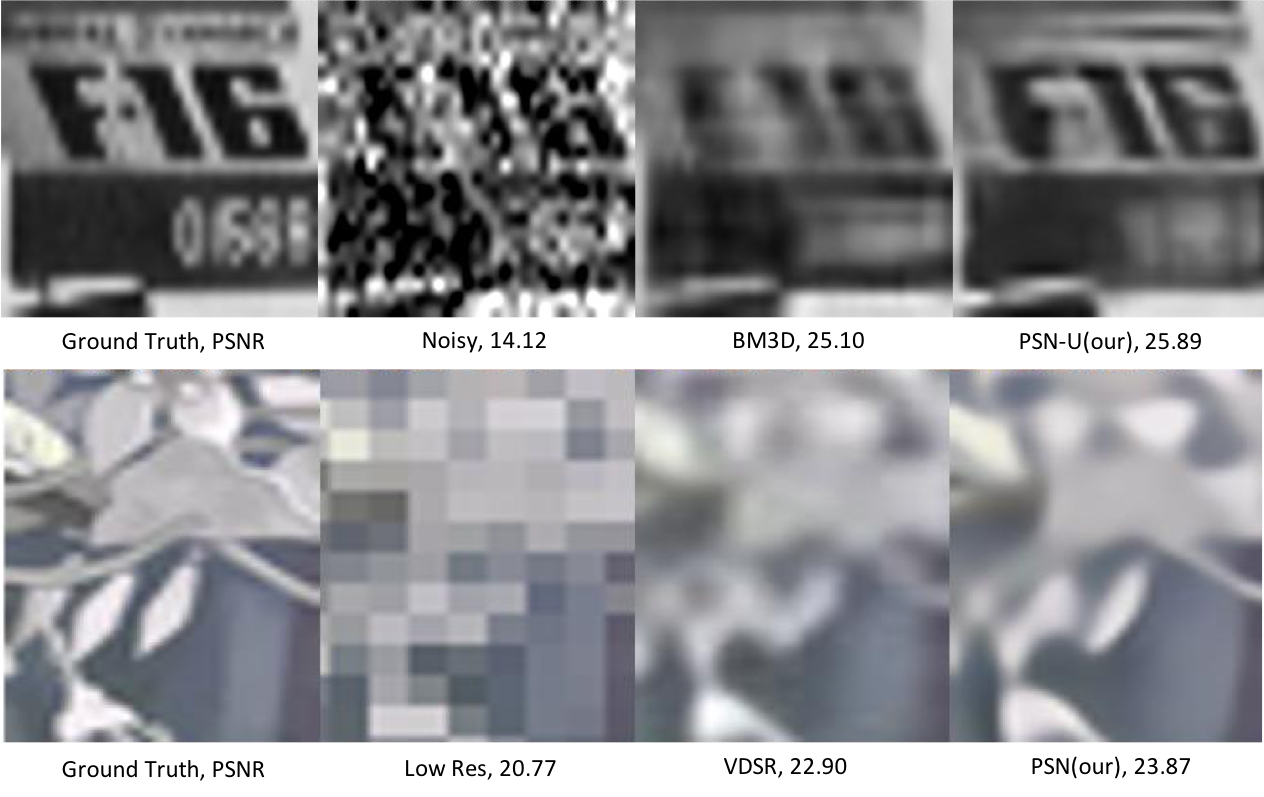}
  \caption{ The first row  shows the de-noising results of “Airpl.” (Set12) compared against BM3D \cite{dabov2007image} with added Gaussian noise ($\sigma=50$). The second row illustrates the super-resolution output of “Comic”(Set14) compared against VDSR \cite{kim2016accurate} with a down-sampling (scale) factor of 4. PSN recovers finer details and more complex structures than the baselines. }
  \label{fig_additional_results}
   %\nvaslpace{-0.5cm}
\end{figure}

\textbf{Results:} From Table.~\ref{table_superres}, it is clear that PSN achieves state-of-the-art results both in terms of PSNR and SSIM for all four benchmarks for all scales by a significant margin. This demonstrates the efficacy of the algorithm in application to the image super-resolution problem. Fig.~\ref{fig_s5} and Fig.~\ref{fig_additional_results} present some qualitative results. Notice that PSN recovers complex structures more clearly. More results are presented in the supplementary.

%%%%%%%%%%%%%%%%%%%%%%%%%%%%%%%%%%%%%%%%%%%%%%%%%%%%%%%%%%%%%%%%%%%%%%%%%%%%%%%%%%%%%%%%%%%%%%%%

\begin{table}
\small
\centering
\begin{tabular}{l*{6}{c}r}

  \hline
    \hline
   Algorithm &  Scale & SET5  & SET14 &  BSDS100 & URBAN100 \\
   &  & PSNR / SSIM & PSNR / SSIM &  PSNR / SSIM & PSNR / SSIM \\

                  \hline
                      \hline
   Bicubic &  &  33.69 / 0.931  & 30.25 / 0.870 &   29.57 / 0.844 & 26.89 / 0.841  \\
 %    A+ \cite{timofte2014a+}&  &    36.60 / 0.955  & 32.32 / 0.906  &   31.24 / 0.887 & 29.25 / 0.895  \\
 %       RFL \cite{schulter2015fast} &  &  36.59 / 0.954  & 32.29 / 0.905 &    31.18 / 0.885 & 29.14 / 0.891  \\
  %     SelfExSR \cite{huang2015single}&  &  36.60 / 0.955  &  32.24 / 0.904 &    31.20 / 0.887 & 29.55 / 0.898  \\
  %     SRCNN \cite{dong2016image}& &  36.72 / 0.955  & 32.51 / 0.908 &    31.38 / 0.889 & 29.53 / 0.896  \\
           FSRCNN \cite{dong2016accelerating}&  &  37.05 / 0.956  & 32.66 / 0.909 &     31.53 / 0.892 &  29.88 / 0.902  \\
  %          SCN  \cite{wang2015deep}& 2X & 36.58 / 0.954  &   32.35 / 0.905 &     31.26 / 0.885 &  29.52 / 0.897  \\
               DRCN \cite{kim2016deeply}&  & 37.63 / 0.959 &   33.06 / 0.912 &    31.85 / 0.895 &  30.76 / 0.914  \\
               LapSRN  \cite{lai2017deep}& 2X & 37.52 / 0.959  &   33.08 / 0.913 &   31.80 / 0.895  &   30.41 / 0.910  \\
               DRRN \cite{tai2017image}&  & 37.74 / 0.959  &   33.23 / 0.914 &    32.05 / 0.897 &  31.23 / 0.919  \\
               VDSR \cite{kim2016accurate}&  & 37.53 / 0.959  &   33.05 / 0.913 &    31.90 / 0.896 &   30.77 / 0.914 \\
               \hline
               PSN (Ours) &  &  \textbf{38.09 / 0.960}   & \textbf{33.68 / 0.919}  &  \textbf{32.33 / 0.901}    &  \textbf{31.97 / 0.921}   \\
  \hline
  \hline  
    Bicubic &  &  30.41 / 0.869  & 27.55 / 0.775 &   27.22 / 0.741 & 24.47 / 0.737 \\
 %    A+ \cite{timofte2014a+}&  &    32.62 / 0.909  & 29.15 / 0.820 &   28.31 / 0.785 &  26.05 / 0.799 \\
 %       RFL \cite{schulter2015fast}&  &  32.47 / 0.906  & 29.07 / 0.818 &    28.23 / 0.782 & 25.88 / 0.792  \\
 %      SelfExSR \cite{huang2015single}&  &  32.66 / 0.910  &  29.18 / 0.821 &    28.30 / 0.786 & 26.45 / 0.810  \\
  %      SRCNN \cite{dong2016image}& &  32.78 / 0.909  & 29.32 / 0.823 &     28.42 / 0.788 &  26.25 / 0.801  \\
           FSRCNN \cite{dong2016accelerating}&  &   33.18 / 0.914  &  29.37 / 0.824 &     28.53 / 0.791  &  26.43 / 0.808  \\
 %          SCN  \cite{wang2015deep}& 3X & 32.62 / 0.908  &   29.16 / 0.818 &     28.33 / 0.783 &   26.21 / 0.801   \\
               DRCN \cite{kim2016deeply}&  & 33.83 / 0.922 &   29.77 / 0.832 &    28.80 / 0.797 &  27.15 / 0.828  \\
               LapSRN  \cite{lai2017deep}& 3X & 33.82 / 0.922  &   29.87 / 0.832 &    28.82 / 0.798  &   27.07 / 0.828   \\
               DRRN \cite{tai2017image}&  & 34.03 / 0.924   &   29.96 / 0.835 &    28.95 / 0.800 &  \textbf{27.53} / 0.764  \\
               VDSR \cite{kim2016accurate}&  & 33.67 / 0.921   &    29.78 / 0.832 &    28.83 / 0.799 &   27.14 / \textbf{0.829}  \\
                              \hline
               PSN (Ours) &  & \textbf{34.56 / 0.927}  &  \textbf{30.14 / 0.845}   &  \textbf{29.26 / 0.809}    &  {27.43 / 0.757} \\

  \hline
  \hline

    Bicubic &  &  28.43 / 0.811  &  26.01 / 0.704 &   25.97 / 0.670  & 23.15 / 0.660  \\
 
  %   A+ \cite{timofte2014a+}&  &     30.32 / 0.860  &  27.34 / 0.751  &    26.83 / 0.711 & 24.34 / 0.721  \\
  %      RFL \cite{schulter2015fast}&  &  30.17 / 0.855  &  27.24 / 0.747 &    26.76 / 0.708  & 24.20 / 0.712  \\
 %      SelfExSR \cite{huang2015single}&  &  30.34 / 0.862  &  27.41 / 0.753 &    26.84 / 0.713 & 24.83 / 0.740 \\
 %      SRCNN \cite{dong2016image}& &  30.50 / 0.863  & 27.52 / 0.753 &    26.91 / 0.712 & 24.53 / 0.725  \\
           FSRCNN \cite{dong2016accelerating}&  &  30.72 / 0.866  & 27.61 / 0.755 &     26.98 / 0.715 &  24.62 / 0.728  \\
 %           SCN  \cite{wang2015deep}& 4X & 30.41 / 0.863  &    27.39 / 0.751  &     26.88 / 0.711 &  24.52 / 0.726  \\
               DRCN \cite{kim2016deeply}&  & 31.54 / 0.884 &   28.03 / 0.768  &     27.24 / 0.725 &  25.14 / 0.752 \\
               LapSRN  \cite{lai2017deep}& 4X & 31.54 / 0.885  &   28.19 / 0.772 &   27.32 / 0.727  &   25.21 / 0.756 \\
               DRRN \cite{tai2017image}&  & 31.68 / 0.888  &   28.21 / 0.772 &    27.38 / 0.728 &  25.44 / 0.764 \\
               VDSR \cite{kim2016accurate}&  & 31.35 / 0.883  &    28.02 / 0.768 &    27.29 / 0.726 &  25.18 / 0.754 \\                              \hline

               PSN (Ours) &  &  \textbf{32.36 / 0.896}  & \textbf{28.40 /  0.786}    &   \textbf{27.73 / 0.742}   &  \textbf{25.63 / 0.768 } \\

  \hline
      \hline
\end{tabular}

\caption{The PSNR and SSIM results of several algorithms on four image super resolution benchmarks. PSN outperforms all previous algorithms significantly and consistently (except for the 3X case on Urban100). PSN also outperforms the works of \cite{timofte2014a+,schulter2015fast,huang2015single,dong2016image,wang2015deep} on all four benchmarks, whose specific results we present in the supplementary due to space constraints. }
\label{table_superres}
\end{table}

%%%%%%%%%%%%%%%%%%%%%%%%%%%%%%%%%%%%%%%%%%%%%%%%%%%%%%%%%%%%%%%%%%%%%%%%%%%%%%%%%%%%%%%%%%%%%%%%
 
\subsection{Conclusion}

%\textbf{Conclusion}

We proposed a theoretically motivated novel deep architecture for image recovery, inspired from the half quadratic algorithm and utilizing the proximal operator. Extensive experiments in image denoising and image super resolution demonstrated the proposed Proximal Splitting Network is effective and achieves a new state-of-the-art on both tasks. Furthermore, the proposed framework is flexible and can be potentially applied to other tasks such as image in-painting and compressed sensing, which are left to be explored in future work.

\newpage

% Thus, the output of the network is the high frequency components of an image. While VDSR is inspired by VGG-net which is used for ImageNet classification \cite{simonyan2014very}, we provide an alternative way to design the a architecture of VDSR.

\section{ Appendix: Algorithms as Special Cases of the Proximal Splitting Network Optimization Problem}

\textbf{VDSR \cite{kim2016accurate} as a special case.}  We now describe the relationship of Proximal Splitting Networks to some of the other deep learning approaches.  The authors in \cite{kim2016accurate} present a single-image super-resolution method called VDSR.  In this work, they use a very deep convolutional network with residual-learning. We find that VDSR is special case of our formulation. VDSR can be modelled by modifying Eq.~12 (from the paper). We set $S=1$, $K^T$ to be the bi-cubic up-sampling filter, $\beta$   to be 2 and where $y$ is the low res image and $x^0$ the up-sampled low res image via bi-cubic interpolation. Further, the last convolution layer of $\Gamma^1_\Theta$ is a filter that can be represented by a matrix with weight equivalents to $(I+K^TK)^{-1}$ and the loss objective being the $l_2$ loss, we will have the following optimization problem:
\begin{equation}
\begin{aligned}
  \min_{{\Theta}}\|{x_{gt}} - x_{1}\|^2_2 ~~~~\text{s.t}~~~~   v_1=\Gamma^1_\Theta(x_{0}), ~~~   x_1=v_1+K^Ty
\end{aligned}
\label{eq:14}
\end{equation}
Here, $\Gamma^1_\Theta$ is modelled to be a deep CNN that consists of 20 layers. This formulation is the exact formulation of the VDSR method.

\textbf{DnCNN \cite{zhang2017beyond} as a special case.} Similarly, the Denoising Convolutional Neural Networks (DnCNN \cite{zhang2017beyond}) can also be modelled as a special case of the PSN optimization problem (Eq.~12 from the paper). In this work the authors propose a CNN for image denoising. DnCNN model has the ability to recover images when the noise level is unknown. DnCNN can be represented by the formula in Eq.~\ref{eq:14}  if $\Gamma^1_\Theta$ is a deep convolutional neural network, $x^0$ is the noisy image and  $K^T$ being the identity matrix. The formulation then describes the architecture of DnCNN.

Thus, we find that some previous approaches can be modelled as special cases of our formulation (Eq.~12 from the paper). Our approach we find, not only theoretically generalizes these methods, but also outperforms them practically on two image restoration tasks.

%The depth of $\Gamma^1_\Theta$ could be set to 17 and each layer is consisted of convolutional filters followed by Relu and batch normalization.% and the last convolution layer of $\Gamma^1_\Theta$ is a delta filter(ADD Reference) with weight $\beta^{-1}$. 

Furthermore, deep multi-scale convolutional neural network for dynamic scene deblurring \cite{nah2017deep} is another special case of our approach.
In this work, they proposed a blind deblurring method with CNN. the proposed network is a multi-scale convolutional neural network that recovers sharp images where blur is caused by several motion filters. To show that this approach is special case of our method, we need to present the formula of combing PSN with multi scale architecture first. The optimization formula of this combination is:

\begin{equation}
\begin{aligned}
&  \min_{{\Theta},D,C} \Sigma_i^S\|{M_ix_{gt}}-x_{i}\|_2^2\\
& \text{s.t}\\
&   v_t=\Gamma^t_\Theta(D_t^Tx_{t-1}+C_tM_tK^Ty) \\
&   x_t=v_{t}-2\beta^{-1}[M_t K^T(Kv_{t}-y)], \; t = 1, \ldots, %S.
\end{aligned}
\label{eq:13}
\end{equation}
Where $D^T$ is a de-convolution filter\cite{noh2015learning}. $M_t$ is sub sampling matrix that reduce the size of the vector that is multiplied with. $M_S$ is the identify matrix, $M_{i}$ is a down sampling matrix by $2^{S-i}$.

To show that the proposed approach in \cite{nah2017deep} is special case of PSN ,we need to manipulate with the value of $\beta$ and $K^T$ in Eq.~\ref{eq:13} since  the filter $k$ is unknown. Thus, it can be written as:
\begin{equation}
\begin{aligned}
&  \min_{{\Theta},D^T,C} \Sigma_i^S\|{M_tx_{gt}}-x_{i}\|_2^2\\
& \text{s.t}\\
&   v_t=\Gamma^t_\Theta(D_t^Tx_{t-1}+C_tM_ty) \\
&   x_t=v_{t}+M_ty \; t = 1, \ldots, S.
\end{aligned}
\label{eq:multi}
\end{equation}
Optimizing this function is exactly equivalent to the approach in \cite{nah2017deep} when $\Gamma^t_\Theta$ is a deep residual network \cite{he2016deep}

By applying the same methodology that we used in the previous three cases, we can show that our approach is general method of \cite{lai2017deep,xu2014deep} too.
\section{Proof of Eq.~4 }

In Eq.~3  (from the paper) ,$h(z)$ can be approximated via the second order of Taylor series since it is twice differentiable. Thus, the optimization problem in Eq.~3 (from the paper) will be
\begin{equation}
prox_{h,\beta}(x) = arg  \min_{{z}}\beta\|{z} - {x}\|^2_2 +h(x)+\nabla h(x)^T(z-x)+ \frac{1}{2}(z-x)^T\nabla^2 h(x)(z-x)
\label{eq:ap}
\end{equation}

Eq.~\ref{eq:ap} is convex. Therefore, Minimizing Eq.~\ref{eq:ap} can be found by taking the first derivative and computing the roots(when the function equals zero). The result will be:

\begin{equation}
prox_{h,\beta}(x) = x-{[\nabla^2 h(x)+\beta I/2]}^{-1}\nabla h(x)
\label{eq:ap21}
\end{equation}

when $\beta$ is large, the proximal of function $h$ can be approximated to be
\begin{equation}
prox_{h,\beta}(x) \approx x-2\beta^{-1}\nabla h(x)
\label{eq:ap2}
\end{equation}

\section{Complexity}

Table~\ref{table:time} shows the run times of different methods for denoising. The input images have three different sizes($256\times 256 $, $512\times 512$ and $1024\times 1024 $. We see that the two versions of the PSN network, PSN-K and PSN-U are one of the fastest algorithms with less than 0.1 seconds for images less than $512 \times 512$. Though it is slightly slower than the networks of TNRD \cite{chen2017trainable} and DnCNN \cite{zhang2017beyond}, it still processes faster than 0.5 seconds for a $1024\times 1024$ image.
\begin{table}
\small
\begin{tabularx}{\textwidth}{ |X|X|X|X|X|X|X|X|X|X| }
  \hline
  Method & BM3D \cite{dabov2007image}& WNNM \cite{gu2014weighted}& EPLL \cite{zoran2011learning}& MLP \cite{burger2012image}& CSF \cite{schmidt2014shrinkage}&TNRD \cite{chen2017trainable} & DnCNN \cite{zhang2017beyond}& PSN-K & PSN-U\\
  \hline
   $256\times 256 $ & 0.65 & 203.1  & 25.4 & 1.42 & 2.11 & 0.010 & 0.016 & 0.017 & 0.018 \\
  \hline
    $512\times 512 $& 2.85 & 773.2  & 45.5  & 5.51 &  5.67 &  0.032  & 0.060 & 0.072 & 0.081\\
      \hline
         $1024\times 1024 $ & 11.89 & 2536.4 & 422.1  &  19.4  & 40.8 & 0.116 &  0.235 & 0.345 & 0.378\\
      \hline

\end{tabularx}
\caption{The complexity in seconds for 3 different sizes}
\label{table:time}
\end{table}

\newpage
\section{Complete Tabular result for Super-resolution}

Table.~\ref{table_superres} shows the full version of Table.~3 from the main paper. This is the complete result for the  super-resolution experiments. We find that PSN still achieves state-of-the-art results on most benchmarks and settings.
%%%%%%%%%%%%%%%%%%%%%%%%%%%%%%%%%%%%%%%%%%%%%%%%%%%%%%%%%%%%%%%%%%%%%%%%%%%%%%%%%%%%%%%%%%%%%%%%

\begin{table}
\small
\begin{tabular}{l*{6}{c}r}

  \hline
    \hline
   Algorithm &  Scale & SET5  & SET14 &  BSDS100 & URBAN100 \\
   &  & PSNR / SSIM & PSNR / SSIM &  PSNR / SSIM & PSNR / SSIM \\

                  \hline
                      \hline
   Bicubic &  &  33.69 / 0.931  & 30.25 / 0.870 &   29.57 / 0.844 & 26.89 / 0.841  \\
 
     A+ \cite{timofte2014a+}&  &    36.60 / 0.955  & 32.32 / 0.906  &   31.24 / 0.887 & 29.25 / 0.895  \\
        RFL \cite{schulter2015fast} &  &  36.59 / 0.954  & 32.29 / 0.905 &    31.18 / 0.885 & 29.14 / 0.891  \\
 
       SelfExSR \cite{huang2015single}&  &  36.60 / 0.955  &  32.24 / 0.904 &    31.20 / 0.887 & 29.55 / 0.898  \\
  
         SRCNN \cite{dong2016image}& &  36.72 / 0.955  & 32.51 / 0.908 &    31.38 / 0.889 & 29.53 / 0.896  \\
  
           FSRCNN \cite{dong2016accelerating}&  &  37.05 / 0.956  & 32.66 / 0.909 &     31.53 / 0.892 &  29.88 / 0.902  \\
 
            SCN  \cite{wang2015deep}& 2X & 36.58 / 0.954  &   32.35 / 0.905 &     31.26 / 0.885 &  29.52 / 0.897  \\
 
               DRCN \cite{kim2016deeply}&  & 37.63 / 0.959 &   33.06 / 0.912 &    31.85 / 0.895 &  30.76 / 0.914  \\
               LapSRN  \cite{lai2017deep}&  & 37.52 / 0.959  &   33.08 / 0.913 &   31.80 / 0.895  &   30.41 / 0.910  \\
               DRRN \cite{tai2017image}&  & 37.74 / 0.959  &   33.23 / 0.914 &    32.05 / 0.897 &  31.23 / 0.919  \\
               VDSR \cite{kim2016accurate}&  & 37.53 / 0.959  &   33.05 / 0.913 &    31.90 / 0.896 &   30.77 / 0.914 \\
               \hline
               PSN (Ours) &  &  \textbf{38.09 / 0.960}   & \textbf{33.68 / 0.919}  &  \textbf{32.33 / 0.901}    &  \textbf{31.97 / 0.921}   \\
  \hline
  \hline  
    Bicubic &  &  30.41 / 0.869  & 27.55 / 0.775 &   27.22 / 0.741 & 24.47 / 0.737 \\
 
     A+ \cite{timofte2014a+}&  &    32.62 / 0.909  & 29.15 / 0.820 &   28.31 / 0.785 &  26.05 / 0.799 \\
        RFL \cite{schulter2015fast}&  &  32.47 / 0.906  & 29.07 / 0.818 &    28.23 / 0.782 & 25.88 / 0.792  \\
 
       SelfExSR \cite{huang2015single}&  &  32.66 / 0.910  &  29.18 / 0.821 &    28.30 / 0.786 & 26.45 / 0.810  \\
  
         SRCNN \cite{dong2016image}& &  32.78 / 0.909  & 29.32 / 0.823 &     28.42 / 0.788 &  26.25 / 0.801  \\
  
           FSRCNN \cite{dong2016accelerating}&  &   33.18 / 0.914  &  29.37 / 0.824 &     28.53 / 0.791  &  26.43 / 0.808  \\
 
            SCN  \cite{wang2015deep}& 3X & 32.62 / 0.908  &   29.16 / 0.818 &     28.33 / 0.783 &   26.21 / 0.801   \\
 
               DRCN \cite{kim2016deeply}&  & 33.83 / 0.922 &   29.77 / 0.832 &    28.80 / 0.797 &  27.15 / 0.828  \\
               LapSRN  \cite{lai2017deep}&  & 33.82 / 0.922  &   29.87 / 0.832 &    28.82 / 0.798  &   27.07 / 0.828   \\
               DRRN \cite{tai2017image}&  & 34.03 / 0.924   &   29.96 / 0.835 &    28.95 / 0.800 &  \textbf{27.53} / 0.764  \\
               
               VDSR \cite{kim2016accurate}&  & 33.67 / 0.921   &    29.78 / 0.832 &    28.83 / 0.799 &   27.14 / \textbf{0.829}  \\
                              \hline

               PSN (Ours) &  & \textbf{34.56 / 0.927}  &  \textbf{30.14 / 0.845}   &  \textbf{29.26 / 0.809}    &  {27.43 / 0.757} \\

  \hline
  \hline

    Bicubic &  &  28.43 / 0.811  &  26.01 / 0.704 &   25.97 / 0.670  & 23.15 / 0.660  \\
 
     A+ \cite{timofte2014a+}&  &     30.32 / 0.860  &  27.34 / 0.751  &    26.83 / 0.711 & 24.34 / 0.721  \\
        RFL \cite{schulter2015fast}&  &  30.17 / 0.855  &  27.24 / 0.747 &    26.76 / 0.708  & 24.20 / 0.712  \\
 
       SelfExSR \cite{huang2015single}&  &  30.34 / 0.862  &  27.41 / 0.753 &    26.84 / 0.713 & 24.83 / 0.740 \\
  
         SRCNN \cite{dong2016image}& &  30.50 / 0.863  & 27.52 / 0.753 &    26.91 / 0.712 & 24.53 / 0.725  \\
  
           FSRCNN \cite{dong2016accelerating}&  &  30.72 / 0.866  & 27.61 / 0.755 &     26.98 / 0.715 &  24.62 / 0.728  \\
 
            SCN  \cite{wang2015deep}& 4X & 30.41 / 0.863  &    27.39 / 0.751  &     26.88 / 0.711 &  24.52 / 0.726  \\
 
               DRCN \cite{kim2016deeply}&  & 31.54 / 0.884 &   28.03 / 0.768  &     27.24 / 0.725 &  25.14 / 0.752 \\
               LapSRN  \cite{lai2017deep}&  & 31.54 / 0.885  &   28.19 / 0.772 &   27.32 / 0.727  &   25.21 / 0.756 \\
               DRRN \cite{tai2017image}&  & 31.68 / 0.888  &   28.21 / 0.772 &    27.38 / 0.728 &  25.44 / 0.764 \\
               VDSR \cite{kim2016accurate}&  & 31.35 / 0.883  &    28.02 / 0.768 &    27.29 / 0.726 &  25.18 / 0.754 \\                              \hline

               PSN (Ours) &  &  \textbf{32.36 / 0.896}  & \textbf{28.40 /  0.786}    &   \textbf{27.73 / 0.742}   &  \textbf{25.63 / 0.768 } \\

  \hline
      \hline
\end{tabular}

\caption{The PSNR and SSIM results of several algorithms for Super-res}
\label{table_superres}
\end{table}

{\small
	\bibliography{nips}
	\bibliographystyle{ieee}
}

\end{document}